# Multilayer Graph Approach to Deep Subspace Clustering

Lovro Sindičić[1] and Ivica Kopriva[2]
[1]Elementary School Antun Mihanović, Zagreb, Croatia
[2]Division of Electronics, Ruđer Bošković Institute, Zagreb, Croatia

*Abstract*—Deep subspace clustering (DSC) networks based on self-expressive model learn representation matrix, often implemented in terms of fully connected network, in the embedded space. After the learning is finished, representation matrix is used by spectral clustering module to assign labels to clusters. However, such approach ignores complementary information that exist in other layers of the encoder (including the input data themselves). Herein, we apply selected linear subspace clustering algorithm to learn representation matrices from representations learned by all layers of encoder network including the input data. Afterward, we learn a multilayer graph that in a multi-view like manner integrates information from graph Laplacians of all used layers. That improves further performance of selected DSC network. Furthermore, we also provide formulation of our approach to cluster out-of-sample/test data points. We validate proposed approach on four well-known datasets with two DSC networks as baseline models. In almost all the cases, proposed approach achieved statistically significant improvement in three performance metrics. MATLAB code of proposed algorithm is posted on https://github.com/lovro-sinda/MLG-DSC.

## I. INTRODUCTION

Clustering or partitioning data into disjoint homogeneous groups is one of the fundamental problems in data analysis [1]. It aims to infer structure from data based on similarity between data points. That is relevant to many applied problems such as image segmentation [2], data mining [3], voice recognition [4], etc. Because sample spaces often have arbitrary shape, distance-based algorithms fail to cluster data in the original ambient domain. Moreover, high-dimensionality of the ambient domain deteriorates performance even further. The reason is the phenomenon known as the *course of dimensionality*. Consequently, identification of low-dimensional structure of data in high-dimensional ambient space is one of the fundamental problems in fields of engineering and mathematics [5]. In many applications, data are well represented by multiple subspaces. Representing data as a union of multiple linear subspaces gave rise to linear subspace clustering [6]-[10]. However, in real world data do not necessarily come from linear subspaces. As noted in [11], in case of face image clustering, reflectance is more likely non-Lambertian and the pose of the subject varies often. Thus, it is more likely that faces of one subject lie on nonlinear manifold than on the linear subspace. One way to address such problem is formulation of subspace clustering algorithms in reproducible kernel Hilbert space (RKHS) employing the *kernel trick*, [12]-[14]. There are, however, two unresolved issues with kernel-based methods: (*i*) some linear subspace clustering algorithms, such as robust version of the sparse subspace clustering [8], cannot be kernelized [14]; (*ii*) after many years of research it is still unclear how to choose the kernel function such that empirical data fit kernel-induced RKHS. As an alternative to kernel-based methods in solving nonlinear subspace clustering problems, neural networks for deep subspace clustering (DSC) emerged [11], [15]-[19]. The main motivation for using them to deal with data generated from nonlinear manifolds was merging their powerful representation learning capability with the linear SC algorithms. Learning a proper linear embedding from data themselves overcomes the fundamental limitation of kernel-based methods. Theoretically, this proper linear embedding, i.e. representation at the output of encoder, should comply with the union-of-linear-subspaces model. Therefore, linear self-expressive SC algorithms with guaranteed subspace-preserving property [7]-[10], should be able to cluster data points in the embedded space according to the subspaces they are generated from. Accordingly, after the learning is finished the representation matrix, often implemented in terms fully connected self-expressive layer [11], [17], [19], is used to compute data affinity matrix. Afterward, spectral clustering [20] assigns labels to data points. However, DSC with self-expressive model in embedded space is analyzed in [21] with the conclusion that model formation in many cases is ill posed. Therefore, data in embedded space may not comply with the union-of-linear-subspaces model. Because of that, significant part of claimed performance can be attributed to an *ad hoc* post-processing technique and not to the specific DSC model. To support this statement we point out that DSC network (DSC-Net) in [11] and maximum entropy subspace clustering network (MESC-Net) in [19] set to zero elements of learned representation matrix less than a threshold. Thereby, the threshold value is adjusted to maximize clustering accuracy and that requires external (hard) labels. In unsupervised learning scenario, to which SC belongs, such *ad hoc* post-processing is not acceptable. That is the reason why in comparative performance analysis in Section IV, we reported as baseline results of DSC-Net and MESC-Net those obtained from learned, but not post-processed, representation matrices.

Motivated by outlined reasons, and also inspired by multi-view SC [22], [23], we propose herein a multilayer-graph-based post-processing approach to DSC. When learning of DSC network is completed, we further explore learned representations in all the layers of the encoder part of the network (including the input data themselves). After tuning of selected shallow single view SC algorithm, such as [9], on all the learned representations independently, we compute the multilayer graph [24]. The learned multilayer graph integrates complementary information that exist in network's layers after the learning is finished, but it was ignored previously. To increase robustness to errors we rely on mathematically tractable intra-subspace projection dominance (IPD) property [25], to keep only $d$ dominant coefficients in learned representations matrices, where $d$ stands for the *a priori* known subspace dimension. To further increase robustness to errors, instead of standard normalized graph Laplacians, we use shifted graph Laplacians [26]. To the best of our knowledge, this is the first attempt to apply a multilayer graph approach in a multi-view like fashion to improve performance of a single-view DSC networks.

The rest of the paper is organized as follows. In Section II we revisit a background and related work. We present multilayer graph approach to DSC in Section III. Experimental results are presented in Section IV, while conclusions are given in Section V.

## II. BACKGROUND AND RELATAED WORK

### A. DSC Networks

Let $\mathbf{X}^0 \in \mathbb{R}^{D_0 \times N}$ represent dataset comprised of $N$ samples in $D_0$-dimensional ambient (input) space. Let $\left\{\mathbf{X}^v \in \mathbb{R}^{D_v \times N}\right\}_{v=1}^{L/2}$, stand for outputs of encoder layers, where $L$ denotes the overall number of layers (encoder and decoder), and $\mathbb{R}$ denotes set of real numbers. Let $\left\{\mathbf{X}^v = \mathbf{X}^v \mathbf{C}^v\right\}_{v=0}^{L/2}$ stand for self-representation models, where $\left\{\mathbf{C}^v \in \mathbb{R}^{N \times N}\right\}_{v=0}^{L/2}$ stand for representation matrices. To cluster data generated from nonlinear manifolds DSC networks [11], [16], [18], [19] solve the following optimization problem:

$$\min_{\mathbf{C}^{L/2}, \Theta} \left\|\mathbf{X} - \tilde{\mathbf{X}}_\Theta \mathbf{C}^{L/2}\right\|_F^2 + \lambda_1 f\left(\mathbf{C}^{L/2}\right) + \lambda_2 h\left(\mathbf{X}, \mathbf{X}_\Theta^{L/2}, \mathbf{C}^{L/2}\right)$$

such that: $\mathbf{X}_\Theta^{L/2} = \Phi_e(\mathbf{X}, \Theta)$ and $\text{diag}\left(\mathbf{C}^{L/2}\right) = \mathbf{0}$

(1)

In (1) $\Theta$ stands for network parameters, and $\Phi_e$ denotes network embedding (encoder output) of data $\mathbf{X}$, and $\tilde{\mathbf{X}}_\Theta$ stands for reconstruction of $\mathbf{X}$. $f$ imposes regularization on the representation matrix $\mathbf{C}^{L/2}$, and $h\left(\mathbf{X}, \mathbf{X}_\Theta^{L/2}, \mathbf{C}^{L/2}\right)$ plays critically important role in removing trivial solutions and specifying properties of nonlinear mapping and embedding space. Two constraints were imposed on representation matrix in [11]: $f\left(\mathbf{C}^{L/2}\right) = \left\|\mathbf{C}^{L/2}\right\|_1$ and $f\left(\mathbf{C}^{L/2}\right) = \left\|\mathbf{C}^{L/2}\right\|_2$ leading respectively to DSC-L1 and DSC-L2 networks. The work [19] imposes the entropy constraint on representation matrix: $f\left(\mathbf{C}^{L/2}\right) = \sum_{i=1}^N \sum_{j=1}^N c_{ij}^{L/2} \ln c_{ij}^{L/2}$, such that $\mathbf{C}^{L/2} \geq \mathbf{0}$. In both networks $h\left(\mathbf{X}, \mathbf{X}_\Theta^{L/2}, \mathbf{C}^{L/2}\right) = \left\|\mathbf{X}_\Theta^{L/2} - \mathbf{X}_\Theta^{L/2} \mathbf{C}^{L/2}\right\|_F^2$. By minimizing the self-expressive representation term, the latent representation $\mathbf{X}_{\Theta_e}^{L/2}$ is encouraged to obey the union-of-linear-subspaces structure. Work in [11] was the first to formally replace the self-expressive term in (1) by fully connected layer, i.e. coefficients of $\mathbf{C}^{L/2}$ are formally substituted by the network parameters $\Theta_s$. In other words, we now have $\mathbf{X}_{\Theta_e}^{L/2} = \mathbf{X}_{\Theta_e}^{L/2} \Theta_s$ and $\Theta_e$ stands for parameters of the encoder. The network is now parameterized in terms of $\Theta := \{\Theta_e, \Theta_s, \Theta_d\}$ where $\Theta_d$ stands for decoder parameters. The network (1) is trained in two stages: (*i*) pretraining without self-expressive layer, and (*ii*) fine tuning which includes the self-expressive layer. Learned $\mathbf{C}^{L/2}$ is now used by spectral clustering [20] to assign labels to data. Evidently, information existing in layers preceding the encoder output layer are ignored.

### B. Intra-subspace dominance property for robust SC

In real applications datasets are likely to contain various types of noise and/or errors. Consequently, data are highly probable to lie near the intersection of multiple dependent subspaces. Therefore, data points with different labels are very likely to be connected with the high edge weights, and that degrades performance of the graph-based methods. In [25] an error-correction method was proposed. It is based on mathematically tractable IPD property in the projection (representation) space. IPD says that small coefficients in the representation matrix always correspond to the projections over errors. The effect of errors can be reduced by keeping $\{d_c\}_{c=1}^k$ largest entries in term of absolute values, and zeroing other entries. Here, $d_c$ equals to the dimensionality of the corresponding subspace and $k$ denotes the overall number of clusters. To eliminate yet another hyperparameter we set all subspaces dimensions to be equal $\{d_c = d\}_{c=1}^k$, and use the existing *a priori* knowledge for $d$. As an example, face images of each subject in the YaleB dataset lie approximately in a $d=9$ subspace [27], [8]. Handwritten digits, such as in MNIST dataset, lie approximately in a $d=12$ subspace [28]. For objects such as those belonging to the COIL20 dataset, recommended subspace dimensions is $d=9$, [29]. Thus, we propose to post-process learned representation matrix:

$$\mathbf{C}^v \leftarrow \lceil \mathbf{C}^v \rceil_d \quad v = 0, \ldots, L/2 \qquad (2)$$

where the operator $\lceil \mathbf{C}^{L/2} \rceil_d$ is applied column-wise keeping $d$ largest coefficients in terms of absolute value and setting others to zero. Thus, as opposed to threshold-based postprocessing of $\mathbf{C}^{L/2}$ by DSC-Net and MESC-Net, (2) does not require tuning and use of external labels.

### C. Data affinity matrix and shifted Laplacian

Once trained, representation matrix $\mathbf{C}^{L/2} = \Theta_s$ is used to compute data affinity matrix:

$$\mathbf{W}^{L/2} = \frac{\left|\mathbf{C}^{L/2}\right| + \left|\mathbf{C}^{L/2}\right|^{\mathrm{T}}}{2} \qquad (3)$$

from which normalized graph Laplacian matrix is computed [26]:

$$\mathbf{L}^{L/2} = \mathbf{I} - \left(\mathbf{D}^{L/2}\right)^{-1/2} \mathbf{W}^{L/2} \left(\mathbf{D}^{L/2}\right)^{-1/2} \qquad (4)$$

with elements of diagonal degree matrix: $\mathbf{D}_{ii}^{L/2} = \sum_{j=1}^N \mathbf{W}_{ij}^{L/2}$. Spectral clustering [20] is applied to (4), to assign labels to data points. Thereby, $k$-means algorithm is applied to left eigenvectors corresponding to $k$ smallest eigenvalues of $\mathbf{L}^{L/2}$. Nevertheless, shifted Laplacian [26]:

$$\mathbf{L}_s^{L/2} = 2\mathbf{I} - \mathbf{L}^{L/2} = \mathbf{I} + \left(\mathbf{D}^{L/2}\right)^{-1/2} \mathbf{W}^{L/2} \left(\mathbf{D}^{L/2}\right)^{-1/2} \quad (5)$$

exhibits increased robustness to noise because *k*-means clustering is now applied to eigenvectors corresponding to *k* largest eigenvalues.

It is evident from (4)/(5) that quality of data affinity matrix plays a key role in performance of SC algorithms. While eq.(3) is used dominantly for that purpose, the alternative is given in [30]. Let us denote SVD of $\mathbf{C}^{L/2}$ as $\mathbf{C}^{L/2} = \mathbf{U}^{L/2} \mathbf{\Sigma}^{L/2} \left(\mathbf{V}^{L/2}\right)^{\mathrm{T}}$. Procedure for constructing affinity matrix based on angular information was proposed in [30] for symmetric representation matrix. If representation matrix is not symmetric by construction, we can compute its symmetric version in a manner of (3) [30]:

$$\mathbf{C}^{L/2} \leftarrow \left(\left|\mathbf{C}^{L/2}\right| + \left|\mathbf{C}^{L/2}\right|^{\mathrm{T}}\right)/2 \ .$$

We can now formulate matrix $\mathbf{M}^{L/2} = \mathbf{U}^{L/2} \left(\mathbf{\Sigma}^{L/2}\right)^{1/2}$, and calculate data affinity matrix according to [30]:

$$\left[\mathbf{W}^{L/2}\right]_{ij} = \left(\frac{\langle \mathbf{m}_i^{L/2}, \mathbf{m}_j^{L/2} \rangle}{\left\|\mathbf{m}_i^{L/2}\right\|_2 \left\|\mathbf{m}_j^{L/2}\right\|_2}\right)^\delta \quad (6)$$

where $\{\mathbf{m}_i^{L/2}\}_{i=1}^N$ stand for columns of $\mathbf{M}^{L/2}$ and $\delta > 0$ is a constant. Since it encodes angular information of the manifold, eq. (6) is expected to yield more accurate graph Laplacian matrix (4), respectively shifted graph Laplacians (5), than eq. (3). The role of $\delta$ is to further re-emphasize affinities between data points. For $\delta > 1$ large values will shrink slightly, while small values will go towards zero much faster.

## III. MULTILAYER GRAPH LEARNING FOR DSC

Due to the reasons elaborated previously, the learned latent representation $\mathbf{X}_{\Theta_e}^{L/2}$ may not fully comply with the union-of-linear-subspaces model. Furthermore, there are complementary information in learned representations $\{\mathbf{X}^v\}_{v=1}^{L/2-1}$, as well as in $\mathbf{X}^0$, that are ignored after the learning is finished. Therefore, we propose to further explore all the representations $\{\mathbf{X}^v\}_{v=0}^{L/2}$ at the end of the learning process. To integrate available complementary information into one graph we propose to use the multilayer graph [24], i.e. MLG-DSC.

### A. MLG-DSC for in-sample data

In this section, we formulate MLG-DSC for in-sample (a.k.a. training) data. In other words, MLG-DSC is applied to whole dataset available, as it is common to many shallow single-view SC algorithms [7]-[10] and multi-view SC algorithms [22], [23], as well as to DSC algorithms [11] [15] [19]. First, target subspace represented by orthonormal basis $\mathbf{U} \in \mathbb{R}^{N \times k}$ is estimated by minimizing distance between it and individual subspaces represented by orthonormal bases: $\{\mathbf{U}^v \in \mathbb{R}^{N \times k}\}_{v=0}^{L/2}$. These bases are obtained as eigenvectors of $\left\{\mathbf{L}_s^v = \mathbf{U}_s^v \mathbf{\Sigma}_s^v \left(\mathbf{U}_s^v\right)^{\mathrm{T}}\right\}_{v=0}^{L/2}$ that correspond with *k* smallest eigenvalues (due to shifted Laplacian) with indexes contained in a set `eig_k`, i.e. $\left\{\mathbf{U}_s^v \leftarrow \mathbf{U}_s^v(:,\mathtt{eig\_k})\right\}_{v=0}^{L/2}$. Thereby, shifted Laplacians $\{\mathbf{L}_s^v\}_{v=0}^{L/2-1}$ are computed analogously to (5). As it is shown in [24], see eq.(8), a modified Laplacian that unifies all the modes (layers in our case) is computed as:

$$\mathbf{L}_{\mathrm{mod}} = \sum_{v=0}^{L/2} \mathbf{L}_s^v - \gamma \sum_{v=0}^{L/2} \mathbf{U}_s^v \left(\mathbf{U}_s^v\right)^{\mathrm{T}} \quad (7)$$

where $\gamma > 0$ is a tradeoff parameter. Now, solution of the clustering problem is given in terms $\mathbf{U} \in \mathbb{R}^{N \times k}$, with *k* eigenvectors that correspond to *k* largest eigenvalues of the modified Laplacian (7). After $\mathbf{U}$ is normalized to unit row norm, $\mathbf{U}_{norm}$, *k*-means clustering is applied to it to assign cluster labels to data points, see also Algorithm 2 in [24]. We summarize proposed multilayer graph approach to deep subspace clustering (MLG-DSC) in Algorithm 1.

___

**Algorithm 1**: MLG-DSC
___

**Input**: Data $\{\mathbf{X}^v \in \mathbb{R}^{D_v \times N}\}_{v=0}^{L/2}$, where *L* denotes the overall number of layers, number of clusters *k*, algorithm for single view linear subspace clustering (SVLSC) with corresponding view-dependent sets of hyperparameters $\{\Theta_{SVLSC}^v\}_{v=0}^{L/2}$, $\gamma$ regularization constant in (7), *d* - *a priori* know subspace dimension, $\delta$ - affinity matrix constant in (6).

**Output**: Assigned cluster indicator matrix $\mathbf{F} \in \mathbb{N}_0^{N \times k}$.

**Step 1**: perform layer-wise computations (*v*=0, ..., *L*/2)

$\mathbf{C}^v \leftarrow SVLSC(\mathbf{X}^v, \Theta_{SVLSC}^v)$

$\mathbf{C}^v \leftarrow \lceil \mathbf{C}^v \rceil_d$

$\mathbf{C}^v \leftarrow \left(\left|\mathbf{C}^v\right| + \left|\mathbf{C}^v\right|^{\mathrm{T}}\right)/2$   % symmetrize if necessary

$\mathbf{C}^v = \mathbf{U}^v \mathbf{\Sigma}^v \left(\mathbf{V}^v\right)^{\mathrm{T}}$

$\mathbf{M}^v = \mathbf{U}^v \left(\mathbf{\Sigma}^v\right)^{1/2}$

$\left[\mathbf{W}^v\right]_{ij} = \left(\frac{\langle \mathbf{m}_i^v, \mathbf{m}_j^v \rangle}{\left\|\mathbf{m}_i^v\right\|_2 \left\|\mathbf{m}_j^v\right\|_2}\right)^\delta \quad i,j=1,...,N$

$\mathbf{L}_s^v = \mathbf{I} + \left(\mathbf{D}^v\right)^{-1/2} \mathbf{W}^v \left(\mathbf{D}^v\right)^{-1/2}$

$\mathbf{L}_s^v \rightarrow \mathbf{U}_s^v \mathbf{\Sigma}_s^v \left(\mathbf{U}_s^v\right)^{\mathrm{T}}$

$\mathbf{U}_s^v \leftarrow \mathbf{U}_s^v(:,\mathtt{eig\_k})$

**Step 2**: compute modified Laplacian

$\mathbf{L}_{\mathrm{mod}} = \sum_{v=0}^{L/2} \mathbf{L}_s^v - \gamma \sum_{v=0}^{L/2} \mathbf{U}_s^v \left(\mathbf{U}_s^v\right)^{\mathrm{T}}$ ,

**Step 3**: Compute $\mathbf{U} \in \mathbb{R}^{N \times k}$ comprised of $k$ eigenvectors that correspond to $k$ largest eigenvalues of $\mathbf{L}_{\text{mod}}$.

**Step 4**: Normalize $\mathbf{U}$ to unit row norm $\mathbf{U}_{\text{norm}}$.

**Step 5**: Apply $k$-means clustering to rows of $\mathbf{U}_{\text{norm}}$ to assign clusters labels to data points: $\mathbf{F} \in \mathbb{N}_0^{N \times k}$.

## B. MLG-DSC for out-of-sample data

Many shallow single-view SC algorithms [7]-[10] and multi-view SC algorithms [22], [23], as well as DSC algorithms [11], [15], [19] are incapable of clustering out-of-sample (a.k.a. test) data. In other words, when new data point $\mathbf{x}^0 \in \mathbb{R}^{D_0 \times 1}$ arrives the whole algorithm has to be rerun again on the augmented data set. That hinders application of mentioned algorithms to on-line learning problems or to large-scale datasets. Herein, we derive extension of proposed MLG-DSC method for out-of-sample data. Let test data point $\mathbf{x}^0$ is embedded with the trained encoder into the latent space $\mathbf{x}^{L/2} \leftarrow \Phi_e(\mathbf{x}^0, \Theta_e)$. Based on cluster labels obtained from Algorithm 1, we obtain partitions of in-sample data:

$$\left\{ \mathbf{X}_c^{L/2} \leftarrow \mathbf{X}_c^{L/2} - \underbrace{\left[ \overline{\mathbf{x}}_c^{L/2} \ldots \overline{\mathbf{x}}_c^{L/2} \right]}_{N_c \text{ times}} \right\}_{c=1}^{k} \quad \overline{\mathbf{x}}_c^{L/2} = \frac{1}{N_c} \sum_{n=1}^{N_c} \mathbf{X}_c^{L/2}(n)$$

$$\bigcup_{c=1}^{k} \mathbf{X}_c^{L/2} = \mathbf{X}^{L/2}, \text{ and } \sum_{c=1}^{k} N_c = N. \quad (8)$$

From $\left\{ \mathbf{X}_c^{L/2} = \mathbf{U}_c \mathbf{\Sigma}_c \mathbf{V}_c^{\mathrm{T}} \right\}_{c=1}^{k}$ we estimate orthonormal bases from the first $d$ left singular vectors of partitions, i.e. $\left\{ \mathbf{U}_c \in \mathbb{R}^{D_{L/2} \times d} \right\}_{c=1}^{k}$ [31]. We assign label $\{c\}_{c=1}^{k}$ to the test point $\mathbf{x}^{L/2}$, according to the minimum of a point-to-a-subspace distance criterion [31]:

$$\left[ \pi\left(\mathbf{x}^{L/2}\right) \right]_c = \begin{cases} 1, \text{ if } c = \arg\min_{l \in \{1,\ldots,k\}} \left\| \tilde{\mathbf{x}}^l - \mathbf{U}_l (\mathbf{U}_l)^{\mathrm{T}} \tilde{\mathbf{x}}^l \right\|_2 \\ 0, \text{ otherwise.} \end{cases}$$
(9)

where $\tilde{\mathbf{x}}^l = \mathbf{x}^{L/2} - \overline{\mathbf{x}}_l^{L/2}$.

## IV. EXPERIMENTAL RESULTS

We validated proposed MLG-DSC algorithm on performance improvements of DSC-L2 network [11] and MESC-Net [19] as baselines. We implemented our method in MATLAB, while original codes were used for DSC-L2 network [32] and MESC-Net [33]. MLG-DSC was applied to DSC-L2 net results on ORL [34], Extended Yaleb (EYaleb) [35], and COIL20 [36] datasets, and on MESC-Net results on MNIST dataset [37]. Regarding SVLSC algorithm in Algorithm 1, we use the multivariate generalization of minimax-concave penalty regularization-based low-rank sparse subspace clustering (GMC) algorithm [9]. It has three hyperparameters. We use accuracy (ACC), normalized mutual information (NMI), and $F_1$ score as performance metrics. Regarding affinity matrix related constant in (6), it is set to $\delta=4$ for ORL dataset, to $\delta=2$ for COIL20 and EYaleb datasets, and to $\delta=6$ for MNIST dataset. Subspace dimension $d$ was set to *a priori* values given in section III.B. To tune hyperparameters we randomly generated 10 subsets containing 7, 46, 50, and 50 data samples from ORL, EYaleb, COIL20 and MNIST datasets in respective order. We validated performance metrics on 100 randomly generated in-sample and out-of-sample subsets. In-sample subsets contained the same number of samples as for hyperparameters tuning. Out-of-sample subsets contained 3, 19, 22, and 50 data samples from ORL, EYaleb, COIL20 and MNIST datasets in respective order. In addition to DSC-L2 net/MESC net and MLG-DSC algorithms, we also validated GMC algorithm on the representation learned at the output of encoder network. That is justified by the fact that encoder output is expected to comply with the union-of-linear-subspaces model. We conducted statistical significance analysis by performing Wilcoxon sum rank test. As can be seen in Tables I to IV, MLG-DSC approach achieved statistically significant improvement in three performance metrics relative to baselines in almost all the cases.

TABLE I: Clustering performance on ORL dataset.

| Algorithm | ACC [%] | NMI [%] | $F_1$[%] |
|---|---|---|---|
| MLG | 82.32±2.38 | 91.31±1.05 | 74.38±2.83 |
|  | 78.47±2.73 | 91.33±1.12 | 62.96±4.14 |
| GMC | 80.36±2.51 | 90.13±1.08 | 71.22±2.98 |
|  | 76.69±2.98 | 90.58±1.22 | 60.24±4.36 |
| DSC-L2 | 79.42±2.87 | 89.42±1.22 | 70.14±3.18 |
|  | 76.81±2.88 | 90.46±1.13 | 60.27±3.95 |
| p MLG vs. DSC-L2 | 3.05×10⁻¹² | 1.68×10⁻²⁰ | 9.00×10⁻¹⁷ |
|  | 5.92×10⁻⁵ | 6.83×10⁻⁷ | 8.39×10⁻¹¹ |
| p GMC vs. DSC-L2 | 0.0590 | 3.18×10⁻⁵ | 0.0226 |
|  | 0.7684 | 0.6398 | 0.7881 |
| p MLG vs. GMC | 9.94×10⁻⁹ | 3.91×10⁻¹² | 3.26×10⁻¹² |
|  | 2.69×10⁻⁵ | 5.86×10⁻⁶ | 4.84×10⁻⁶ |

TABLE II: Clustering performance on COIL20 dataset.

| Algorithm | ACC [%] | NMI [%] | $F_1$[%] |
|---|---|---|---|
| MLG | 81.56±2.21 | 89.42±0.24 | 77.26±2.11 |
|  | 81.62±2.22 | 89.44±1.05 | 76.49±2.32 |
| GMC | 81.31±2.53 | 89.28±1.08 | 77.28±2.28 |
|  | 81.40±2.58 | 89.35±1.15 | 76.61±2.48 |
| DSC-L2 | 76.19±1.88 | 85.25±0.82 | 71.48±1.85 |
|  | 76.69±1.95 | 85.45±0.94 | 70.85±1.96 |
| p MLG vs. DSC-L2 | 9.24×10⁻²⁹ | 2.56×10⁻³⁴ | 2.92×10⁻³¹ |
|  | 1.89×10⁻²⁷ | 6.48×10⁻³⁴ | 1.84×10⁻²⁹ |
| p GMC vs. DSC-L2 | 2.57×10⁻²⁶ | 1.33×10⁻³³ | 6.50×10⁻³¹ |
|  | 3.48×10⁻²⁵ | 1.08×10⁻³³ | 2.24×10⁻²⁹ |
| p MLG vs. GMC | 0.6052 | 0.3679 | 0.9542 |
|  | 0.6767 | 0.8594 | 0.6743 |

TABLE III: Clustering performance on EYaleb dataset.

| Algorithm | ACC [%] | NMI [%] | $F_1$[%] |
|---|---|---|---|
| MLG | 92.47±1.73 | 93.47±0.65 | 87.02±1.71 |
|  | 90.55±1.75 | 92.04±0.87 | 82.94±2.03 |
| GMC | 89.51±1.27 | 92.49±0.59 | 85.37±1.30 |
|  | 87.84±1.39 | 91.38±0.86 | 81.55±1.98 |
| DSC-L2 | 87.91±1.74 | 92.50±0.64 | 84.94±1.50 |
|  | 85.75±1.83 | 90.67±0.95 | 80.11±1.98 |
| p MLG vs. DSC-L2 | 4.26×10⁻³⁰ | 5.81×10⁻¹⁸ | 1.09×10⁻¹⁴ |
|  | 1.33×10⁻³⁰ | 8.53×10⁻¹⁸ | 5.53×10⁻¹⁷ |
| p GMC vs. DSC-L2 | 8.28×10⁻⁹ | 0.8079 | 0.0357 |
|  | 2.08×10⁻¹⁴ | 6.49×10⁻⁷ | 1.43×10⁻⁶ |
| p MLG vs. GMC | 5.38×10⁻²³ | 6.95×10⁻¹⁹ | 1.47×10⁻¹¹ |
|  | 1.62×10⁻²⁰ | 8.77×10⁻⁷ | 3.76×10⁻⁶ |

TABLE IV: Clustering performance on MNIST dataset.

| Algorithm | ACC [%] | NMI [%] | $F_1$[%] |
|---|---|---|---|
| MLG | 65.85±3.84<br>65.49±3.83 | 61.74±2.85<br>61.70±2.92 | 54.39±3.30<br>54.11±3.41 |
| GMC | 65.19±4.48<br>64.96±4.57 | 61.12±3.09<br>61.13±3.13 | 53.78±3.56<br>53.53±3.72 |
| MESC | 63.07±3.23<br>60.99±371 | 74.90±1.68<br>61.03±2.91 | 58.75±2.79<br>51.06±3.50 |
| p MLG vs. MESC | $2.65 \times 10^{-7}$<br>$4.22 \times 10^{-13}$ | $2.56 \times 10^{-34}$<br>0.0763 | $2.16 \times 10^{-17}$<br>$1.31 \times 10^{-8}$ |
| p GMC vs. MESC | $1.03 \times 10^{-4}$<br>$2.10 \times 10^{-10}$ | $2.72 \times 10^{-34}$<br>0.7259 | $2.06 \times 10^{-19}$<br>$2.97 \times 10^{-6}$ |
| p MLG vs. GMC | 0.2532<br>0.5707 | 0.1619<br>0.1671 | 0.1345<br>0.2985 |

## V. CONCLUSION

We proposed multilayer graph approach to integrate information in DSC network after learning. Comparative performance analysis conducted on four datasets against two DSC networks as baselines, demonstrated statistically significant improvement in terms of three clustering metrics.

## ACKNOWLEDGMENT

This work was supported by the Croatian science foundation grant IP-2022-10-6403.